\documentclass[conference]{IEEEtran}
\usepackage[utf8]{inputenc}
\usepackage{esvect}
\usepackage{amsmath}
\usepackage{mathtools}
\usepackage{stfloats}
\usepackage{color}
\usepackage{fixltx2e}
\usepackage{xspace}
\usepackage[nolist,nohyperlinks]{acronym}
\usepackage[font=footnotesize]{subfig}

\title{PCA-Initialized Deep Neural Networks Applied To Document Image Analysis}

\author{
  \IEEEauthorblockN{
    Mathias Seuret\IEEEauthorrefmark{1},
    Michele Alberti\IEEEauthorrefmark{1},
    Rolf Ingold\IEEEauthorrefmark{1},
    and Marcus Liwicki\IEEEauthorrefmark{1}\IEEEauthorrefmark{3},
  }

\IEEEauthorblockA{
  \IEEEauthorrefmark{1}University of Fribourg, Department of Informatics\\
  Bd. de P\'erolles 90, 1700 Fribourg, Switzerland \\
  Email: firstname.lastname@unifr.ch
}

\IEEEauthorblockA{\IEEEauthorrefmark{3} University of Kaiserslautern, Germany}
}

\newacro{dia}[\textsc{Dia\xspace}]{Document Image Analysis}

\newacro{dnn}[\textsc{Dnn\xspace}]{Deep Neural Network}

\newacro{rbm}[\textsc{Rbm\xspace}]{Restricted Boltzmann Machine}

\newacro{cnn}[\textsc{Cnn\xspace}]{Convolutional Neural Network}

\newacro{pca}[\textsc{Pca\xspace}]{Principal Component Analysis}

\newacro{lda}[\textsc{Lda\xspace}]{Linear Discriminant Analysis}

\newacro{rbe}[\textsc{Rbe\xspace}]{Relative Backpropagated Error}

\newacro{ae}[\textsc{Ae\xspace}]{Auto-Encoder}
\newacro{cae}[\textsc{Cae\xspace}]{Convolutional Auto-Encoder}
\newacroplural{cae}[\textsc{Cae}s]{Convolutional Auto-Encoders}
\newacro{mlf}[\textsc{Mlf\xspace}]{Multilevel Features}
\newacro{tlf}[\textsc{Tlf\xspace}]{Top-level Features}
\newacro{gt}[\textsc{Gt\xspace}]{Ground Truth}

\newacro{cmf}[\textsc{Cmf\xspace}]{Central Multilayer Feature}
\newacroplural{cmf}[\textsc{Cmf\xspace}s]{Central Multilayer Features}

\newacro{tlf}[\textsc{lfF\xspace}]{Top-Layer Feature}
\newacroplural{tlf}[\textsc{Tlf\xspace}s]{Top-Layer Features}

\newacro{ftf}[\textsc{Ftf\xspace}]{Fine-Tuned Feature}
\newacroplural{ftf}[\textsc{Ftf\xspace}s]{Fine-Tuned Features}

\newacro{svm}[\textsc{Svm\xspace}]{Support Vector Machine}
\newacroplural{svm}[\textsc{Svm\xspace}s]{Support Vector Machines}

\begin{document}

\maketitle

\begin{abstract}
In this paper, we present a novel approach for initializing deep neural networks, i.e., by turning \ac{pca} into neural layers.
Usually, the initialization of the weights of a deep neural network is done in one of the three following ways: 1) with random values, 2) layer-wise, usually as Deep Belief Network or as auto-encoder, and 3) re-use of layers from another network (transfer learning).
Therefore, typically, many training epochs are needed before meaningful weights are learned, or a rather similar dataset is required for seeding a fine-tuning of transfer learning. 
In this paper, we describe how to turn a \ac{pca} into an auto-encoder, by generating an encoder layer of the \ac{pca} parameters and furthermore adding a decoding layer.
We analyze the initialization technique on real documents. 
First, we show that a \ac{pca}-based initialization is quick and leads to a very stable initialization.
Furthermore, for the task of layout analysis we investigate the effectiveness of \ac{pca}-based initialization and show that it outperforms state-of-the-art random weight initialization methods.
\end{abstract}

\section{Introduction}
A recent trend in machine learning is training very deep neural networks.
Although artificial neurons have been around for a long time, the depth of commonly used artificial neural networks has started to increase significantly only for roughly 15 years~\cite{mit-book}\footnote{Note that deep neural architectures where proposed already much earlier, but they have not been heavily used in practice~\cite{888}}.
This is due to both a combination of higher computational power available to researchers, and to a coming back of layer-wise training methods~\cite{ballard1987modular} (referred to as Deep Belief Networks~\cite{hinton2006fast}, and often composed of Restricted Boltzmann Machines~\cite{smolensky1986}).

However, the training of deep neural networks still suffers from two major drawbacks.
First, training a \ac{dnn} still takes some time, and second, initializing the weights of \ac{dnn} with random values adds some randomness to the obtained results.

Especially for large input images or patches with diverse content, the training time becomes a crucial issue, as many weights have to be trained and a training epoch takes much time.
Historical \ac{dia} is en example for such tasks as input documents could be of several megabytes and the historical documents can be quite diverse.
The diversity has several origins: degradations present in the documents, complexity and variability of the layouts, overlapping of contents, writing styles, bleed-through, etc.

In this paper, we present a novel initialization method based on the \ac{pca} which allows to quickly initialize the weights of a neural network with non-random values, and additionally leads to a high similarity in the weights of two networks initialized on different random samples from a same dataset.

\section{State of the Art}\label{sct:state-of-art}
There are currently three main trends for neural network initialization: 1) random initial initial weights~\cite{glorot2010understanding}, layer-wise unsupervised pre-training~\cite{ballard1987modular,hinton2006fast}, or transfer learning~\cite{caruana1998multitask}.

Random initialization is fast and simple to implement.
The most used one, presented by Glorot \textit{et al.} recommend~\cite{glorot2010understanding} to initialize weights of a neuron in $\left[-\sqrt{N}, \sqrt{N}\right]$, where $N$ is the number of inputs of the neuron.
This initialization is often called ``Xavier'' initialization, motivated by the first author's name in~\cite{glorot2010understanding}.

The link between \ac{pca} and neural networks is not novel.
In 1982, Oja introduced a linear neuron model learning to generate the principal component of its input data~\cite{oja1982simplified}.
Oja also later investigated the possibility of having linear neural layers learn principal and minor components of a \ac{pca}~\cite{oja1992principal}.
Since then, other publication present methods teaching networks to compute \ac{pca}.
\ac{pca} is also used for dimensionality reduction as a pre-processing steps~\cite{ngiam2011multimodal,sun2014deep}.
Chan \textit{et al.} showed that \ac{pca} filters can be convolved and stacked in order to create efficient feature extractors~\cite{chan2015pcanet}.

Note that while we were conducting our experiments (published in a Master thesis~\cite{alberti2016}) a similar, but independent, research has been published at ICLR 2016~\cite{krahenbuhl2015data}.
While the general initial idea of using \ac{pca} for initialization is similar, it is applied to \ac{cnn} in~\cite{krahenbuhl2015data}.
In this work, we introduce a mathematical framework for generating \ac{cae} out of the \ac{pca}, taking into account the bias of neural layers, and provide a deeper analysis of the behavior of \ac{pca}-initialized networks -- both for \ac{cae} and \ac{cnn} -- with a focus on historical document images.

\section{PCA as Auto-Encoder}\label{sct:methodology}
An \ac{ae} is a machine learning method which is trained to encode its input as a (usually) compressed representation, and decode this representation to recover the input with an accuracy as good as possible.
This compressed representation, also called feature vector, contains a representation of the input of the \ac{ae} and can be used for classification purpose.
A \ac{dnn} can be trained by stacking \acp{ae}, as the layers can be trained one after another, without time-consuming back-propagation through all layers.

\subsection{Encoder Computation}

A standard artificial neural layer takes as input a vector $\vec{x}$, multiplies it by a weight matrix $W$, adds a bias vector $\vec{b}$, and applies a non-linear activation function $f$ to the result to obtain the output $\vec{y}$:

\begin{equation}\label{eq:layer}
  \vec{y}_{nn} = f\left(\vec{b} + W\cdot \vec{x}\right)
\end{equation}

An \ac{ae} is usually composed of two neural layers, one which is trained to encode, and the second to decode.
The weights of these layers can have a relationship (e.g., in \acp{rbm}, the matrix of the decoder is the transpose of the matrix of the encoder), but this is not mandatory.

A \ac{pca} is computed by subtracting the mean vector from the data $\vec{m}$ to an input vector $\vec{x}$, and multiplying the result by a matrix $R$ to obtain the result $y$:

\begin{equation}\label{eq:pca}
  \vec{y}_{pca} = R \cdot \left(\vec{x} - \vec{m}\right) = R\cdot \vec{x} - R\cdot \vec{m}
\end{equation}

In order to have a neural layer having a similar behavior as the \ac{pca}, we have to also apply the activation function $f$ to the \ac{pca}, and thus obtain what we call activated \ac{pca}:

\begin{equation}\label{eq:apca}
  \vec{y}_{apca} = f\left(R\cdot \vec{x} - R\cdot \vec{m}\right)
\end{equation}

We want that $\vec{y}_{apca} = \vec{y}_{nn}$.
Let $\vec{x}_0 = \vec{0}$.
Then, we have

\begin{equation}
  f\left(R\cdot \vec{0} - R\cdot \vec{m}\right) = f\left(\vec{b}+W\cdot \vec{0}\right)
\end{equation}

Assuming that $f$ is strictly monotonous, we get that $\vec{b} = -R\cdot \vec{m}$.
We can inject this result in Equation~\ref{eq:apca} to get

\begin{equation}
  f\left(R\cdot \vec{x} + \vec{b}\right) = f\left(\vec{b} + W\cdot \vec{x}\right)
\end{equation}

Thus, we get that $W = R$.
This allows us to set the weights and bias of a neural layer such that it behaves exactly as an activated \ac{pca}.
The additional computational cost for computing $\vec{b}$ is negligible, thus the initialization cost of the network depends only on the number of samples used for computing the \ac{pca}.\footnote{note that the samples used for \ac{pca} can be less than the full training set. We investigate this aspect in the experiments}

In order to decrease the dimensionality of the encoded values, only some of the principal components can be kept.
For reasons of notation simplicity, we will however ignore this in the formulas.
We will also show only the 9 first components in the figures, as having more would make them tiny and harder to analyze.

If the activation function is not strictly monotonous, e.g., as the rectified linear unit, then for each sample the activation probability of each feature is 50\%.
If $N$ features are used, then $\frac{1}{2^N}$ of the samples will not activate any feature.
If this is an issue, this can be tackled by duplicating one of the component and multiplying its weights by -1.

\subsection{Decoder Approximation}
In the previous step, we proved that a \ac{pca} can be used to initialize the weights of a single neural layer.
However, a layer in an \ac{ae} is composed of two neural layers: the encoder and the decoder.
While \ac{pca}-initialized neural layers can be used for generating the encoder, initializing the decoder requires another approach if we want to have two neural layers having the same implementation.
Encoding and decoding an input $\vec{x}$ with two layers, having respectively the weights $W_1$ and $W_2$, and the bias $\vec{b}_1$ and $\vec{b}_2$ is done as the following:

\begin{equation}
  \vec{x} \approx f\left(\vec{b}_2 + W_2 \cdot f\left(\vec{b}_1 + W_1 \cdot \vec{x}\right)\right)
\end{equation}

However, the reconstruction of $\vec{x}$ cannot be perfect for most activation functions $f$.
Thus, we propose to compute $\vec{b}_2$ and $W_2$ by minimizing the squared error of the reconstruction of at least as many samples as there are dimensions in the \ac{pca}.

Let us assume that $X$ is a matrix which columns are composed of at least as many input vectors as components from the \ac{pca} have been kept.
We can create the matrices $X^+$ by appending a row composed of $1$'s to $X$, and $W_1^+$ by appending a column composed of the vector $\vec{b}_1$ to $W_1$.

Then, we can compute the matrix $Y$ storing the encoded samples in its columns as

\begin{equation}
  Y = f\left(W_1^+ \cdot X^+\right)
\end{equation}

Then, to compute $\vec{b}_2$ and $W_2$, we have to find the matrix $W_2^+$ approximating the following with the smallest square error:

\begin{equation}
  X \approx f\left(W_2^+ \cdot Y\right)
\end{equation}

\noindent where $X$ is an input vector filled either with real samples, or random values.
In other words, we have to find the least-squares solution to the following equation system: $W_2^+\cdot Y = f^{-1}\left(X\right)$.
The vector $\vec{b}_2$ is then the last column of $W_2^+$, and $W_2$ is the other columns of $W_2^+$.

\subsection{Differences Between PCA and \acp{ae}}
\label{sct:diff}
The matrix $R$ of a \ac{pca} can be inverted, thus we have

\begin{equation}
  \vec{x} = R^{-1} \cdot R\cdot \left(\vec{x}-\vec{m}\right) + \vec{m}
\end{equation}

\begin{figure}[ht]
  \centering
  \subfloat[\ac{pca} features]{\label{fig:pca-vs-nn-a} \includegraphics[width=.48\columnwidth]{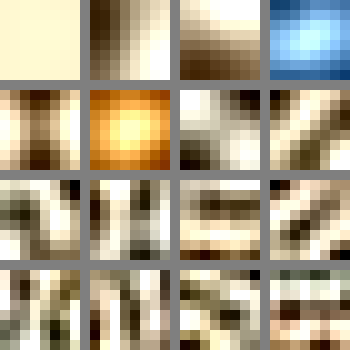}}
  \hfill
  \subfloat[\ac{ae} features]{\label{fig:pca-vs-nn-b} \includegraphics[width=.48\columnwidth]{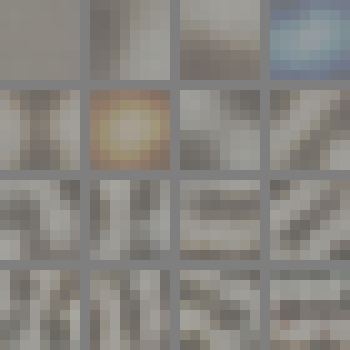}}
  \caption{
    Features computed by a \ac{pca}, and their version obtained when turning the \ac{pca} into an \ac{ae}.
  }
  \label{fig:pca-vs-nn}
\end{figure}

However, in the case of an \ac{ae}, the non-linearity prevents from reconstructing perfectly the input.
In practice, this leads to having a lower contrast when decoding, as the encoded values, which are unbounded when applying a \ac{pca} are restricted to the interval $\left]-1, 1\right[$ when applying a \ac{pca}-initialized neural layer.
This effect is illustrated in Figure~\ref{fig:pca-vs-nn}.

Note that \acp{ae} are used for initializing the weights of neural networks, so the decoding layer is not used for classification tasks.
Thus, this is not an issue for classification if we get low-contrast reconstructions.
However, in order to allow a better visual inspection of the features, we will enhance their contrast in this paper.

\section{Impact Evaluation}\label{sct:experiments}
In this section, we evaluate two aspect of a \ac{pca}-based initialization: first, how stable the features are, and second, what the impact of a \ac{pca}-based initialization is for classification tasks.
All time measurements have been done on a computer with an i5-4570 processor; note that some of the tasks were done too fast for it to increase its frequency.

We used six manuscripts coming from two datasets.
The first dataset is composed of pages from the manuscripts SG857\footnote{St. Gallen, Stiftsbibliothek, Cod. Sang. 857 (13th century)}, SG562\footnote{St. Gallen, Stiftsbibliothek, Cod. Sang. 562 (9th century)}, and GW10\footnote{G. W. Papers, Library of Congress, (18th century)}.
The layout has been annotated with hand-drawn polygons using \textsc{Divadia}~\cite{chen-gt}.
The different classes are out-of-page, background, text, comments, and annotations.
The second dataset is called DIVA-HisDB~\cite{simistira1656a}.
It is composed of pages from the manuscripts CB55\footnote{K\"oln, Fondation Bodmer, Cod. Bodm. 55 (14th century)}, CS18\footnote{St. Gallen, Stiftsbibliothek, Cod. Sang. 18 (11th century)}, and CS863\footnote{St. Gallen, Stiftsbibliothek, Cod. Sang. 863}.
The layout has been annotated using GraphManuscribble~\cite{garz16creating} using a mix of pen-based interactions and automatic polygon generation.

\subsection{Initialization Stability}
\begin{figure}[tb]
  \centering
  \subfloat[CB55, 500 samples]{\label{fig:pca-stab-500} \includegraphics[width=.47\columnwidth]{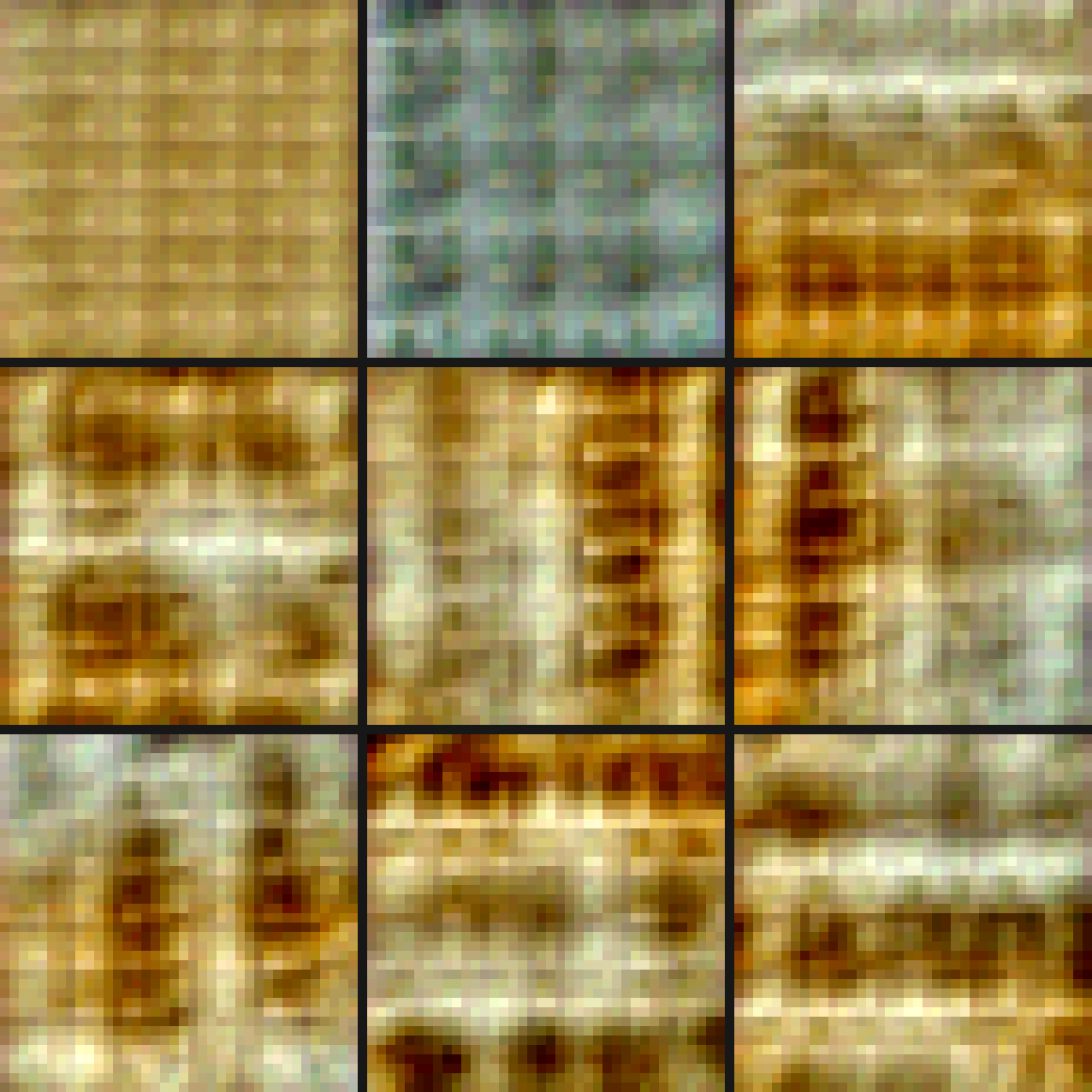}}
  \hfill
  \subfloat[CB55, 5'000 samples]{\label{fig:pca-stab-5000} \includegraphics[width=.47\columnwidth]{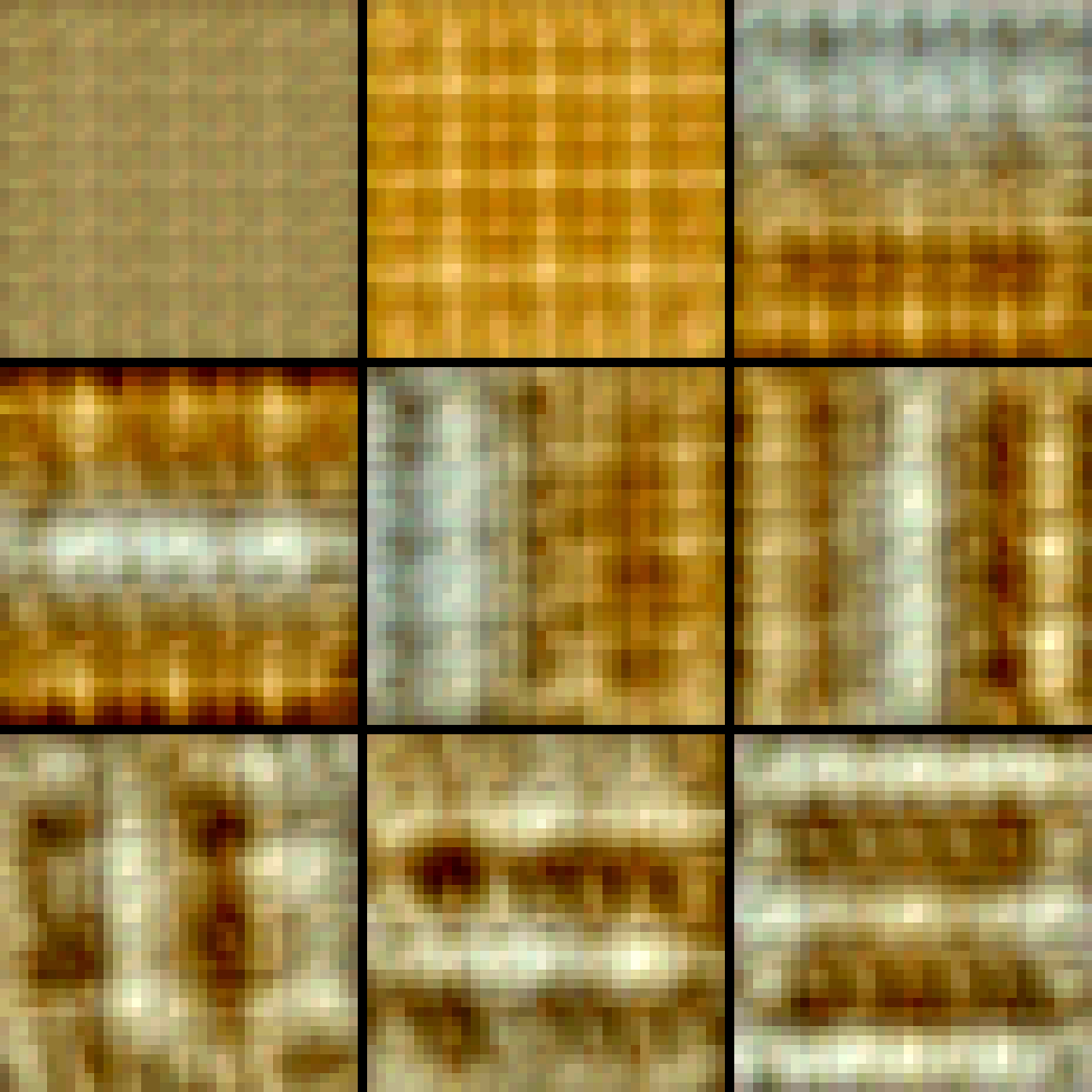}}
  \\
  \subfloat[CB55, 10'000 samples]{\label{fig:pca-stab-10000} \includegraphics[width=.47\columnwidth]{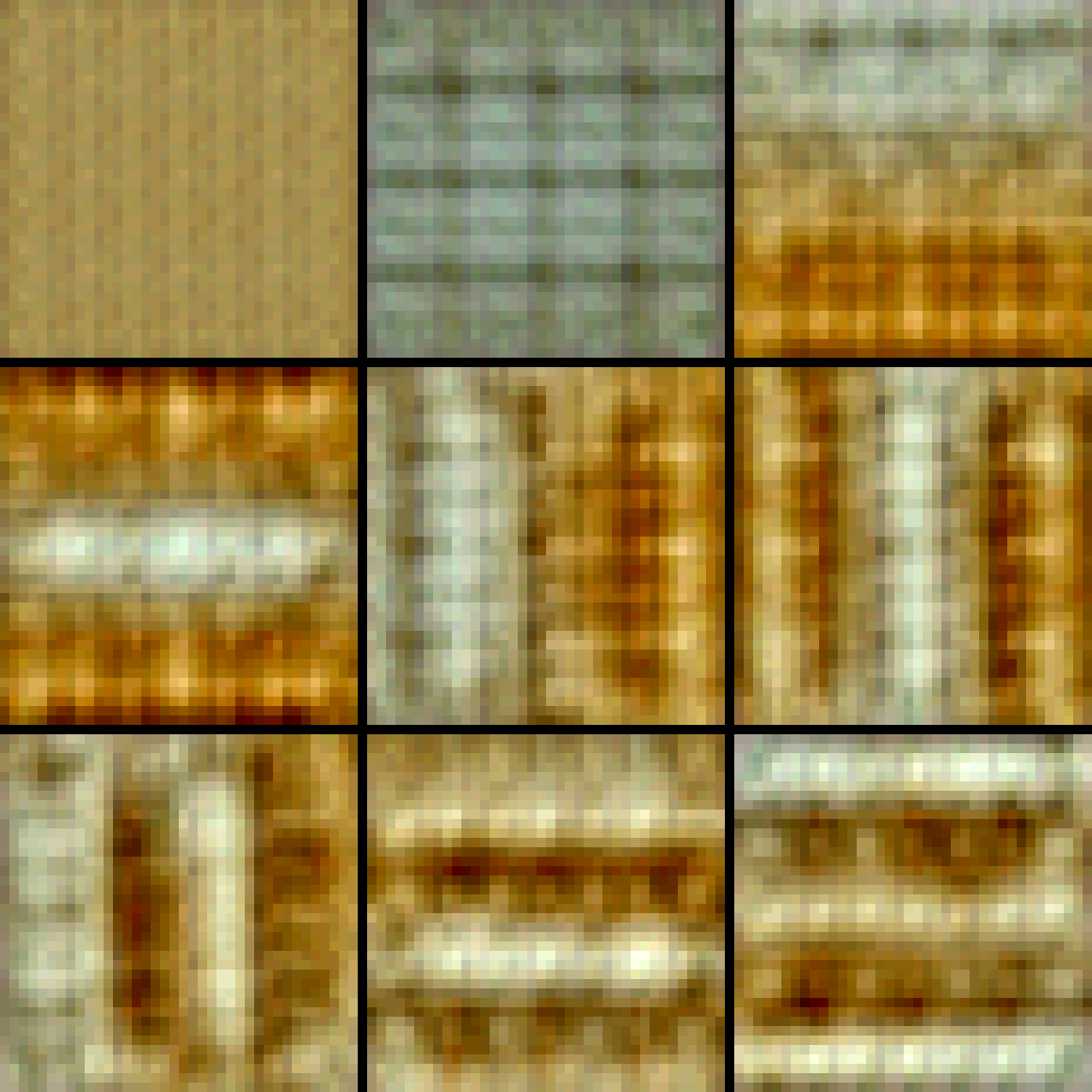}}
  \hfill
  \subfloat[CB55, 50'000 samples]{\label{fig:pca-stab-50000} \includegraphics[width=.47\columnwidth]{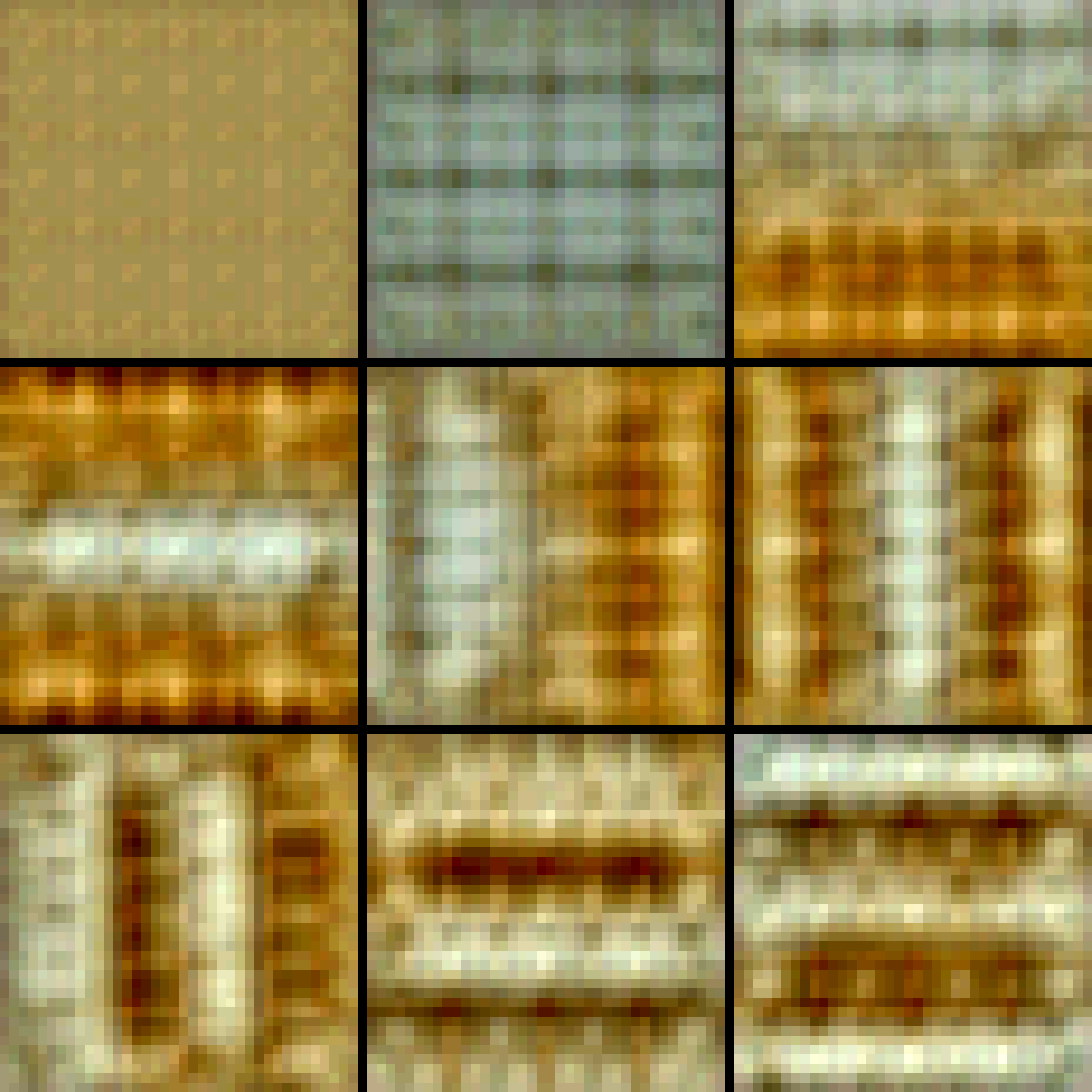}}
  \\
  \subfloat[CS18, page 105]{\label{fig:pca-stab-a} \includegraphics[width=.47\columnwidth]{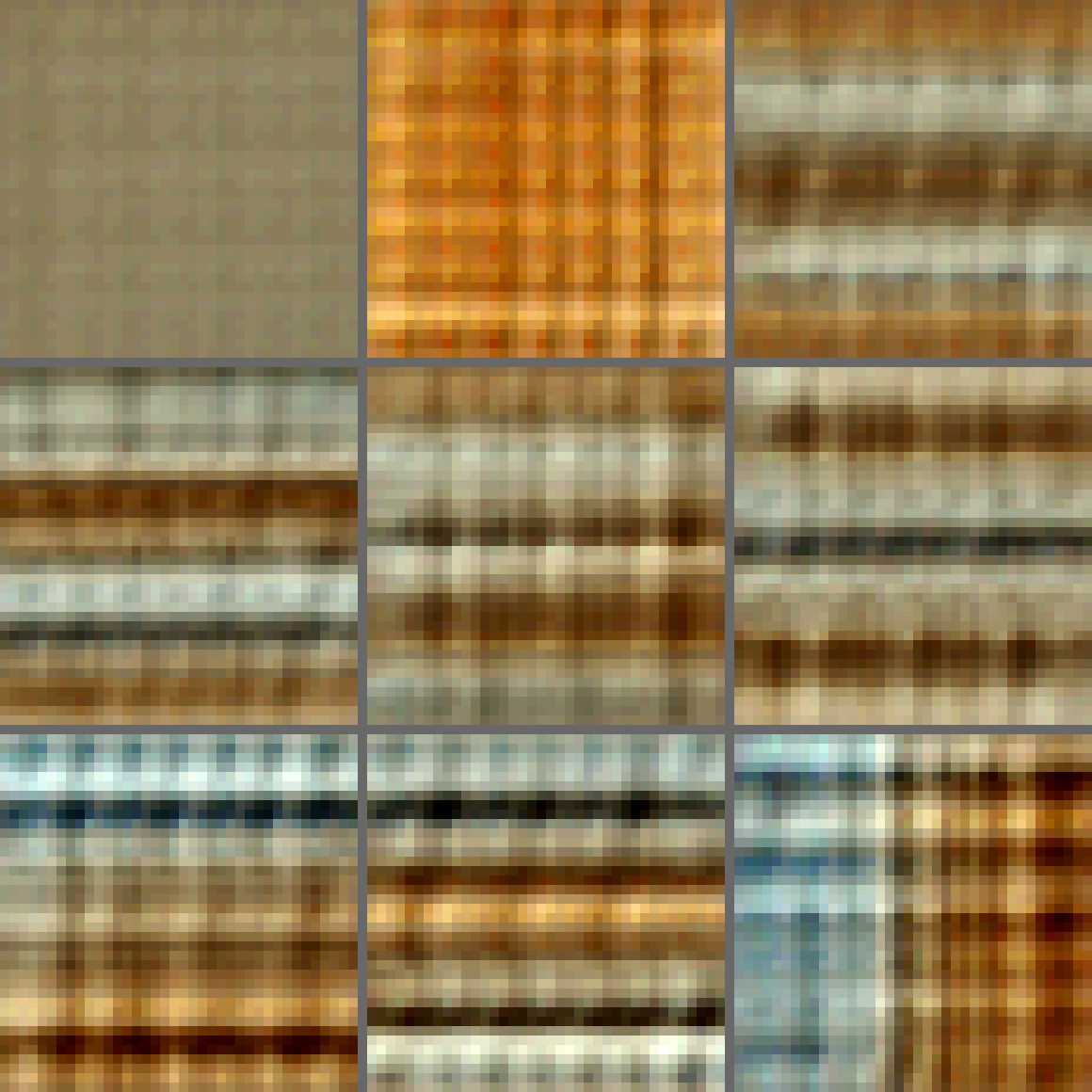}}
  \hfill
  \subfloat[CS18, page 118]{\label{fig:pca-stab-b} \includegraphics[width=.47\columnwidth]{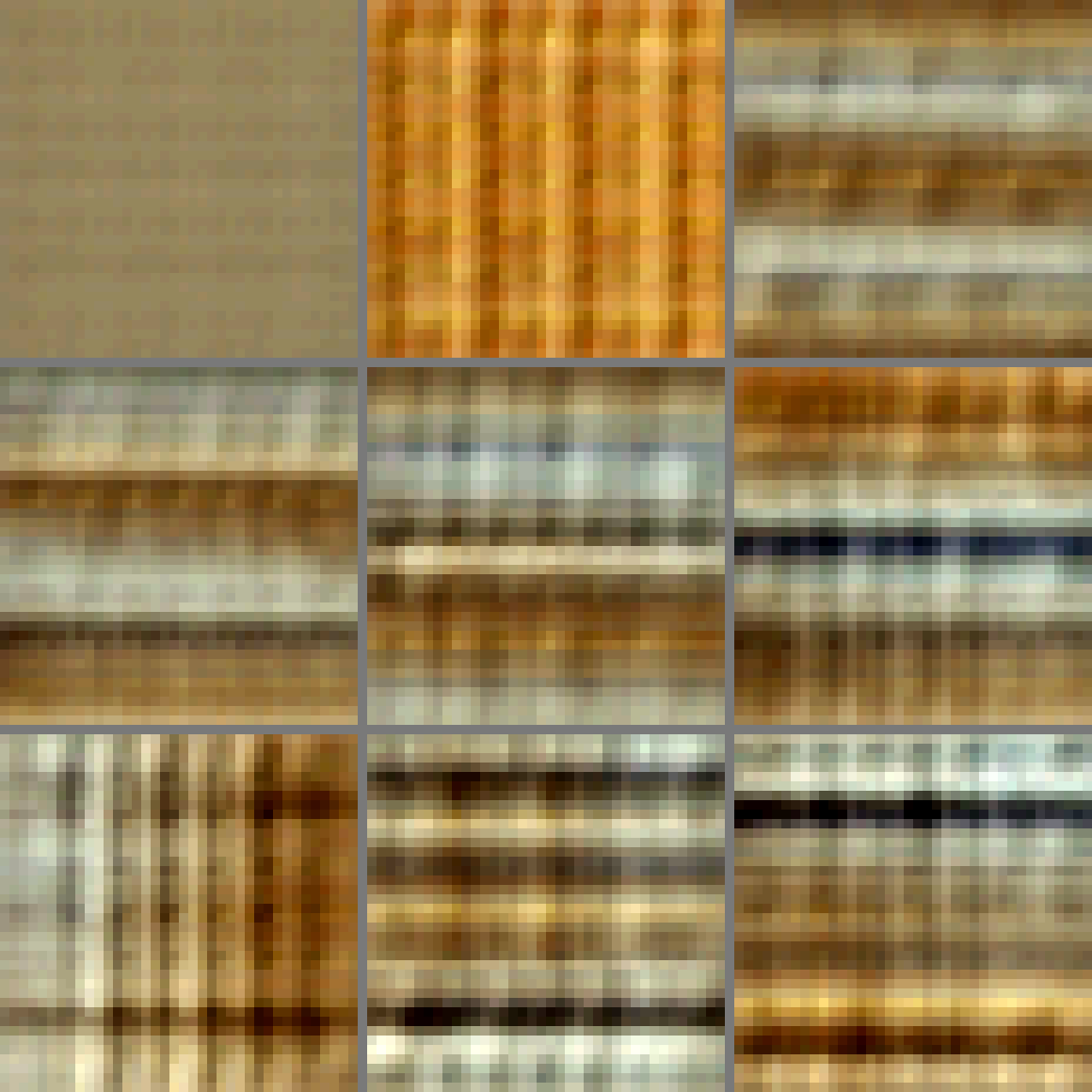}} \\
  \caption{
    Illustration of the stability of features of the third layer of an \ac{ae}.
    The features a, b, c, and d were obtained from different number of samples randomly selected from CB55.
    The features shown in (e) and (f) were obtained using two different pages from a same manuscript.
		Note that the features have been enhanced for better visual inspections (See end of Section~\ref{sct:diff}).
  }
  \label{fig:pca-stability}
\end{figure}

As we keep only some of the principal components of the data samples used for training, we can expect the features to be stable because they will represent main trends of the data and thus be rather stable with regard to the number of samples used for computing the \ac{pca}, or the pages used as sample sources.
Features obtained through \ac{pca}-initialization of three-levels \acp{ae} are shown in Fig.~\ref{fig:pca-stability}.
Note that the six \acp{ae} used for generating the features of Fig~\ref{fig:pca-stability} were trained completely independently.
Differences between initializations with 500 and 50'000 samples lead to similar features, however the initialization times were very different, with less than 1 second for 500 samples, and $\approx2100$ seconds for 50'000 samples.
Note that due to the differences in training and tests, the results we obtained cannot be compared to our previous works.

Another kind of stability can be observed.
If we initialize an \ac{ae} on two different pages of a same manuscript, we can expect the features to be similar.
While it is not the case when training a network as a standard \ac{ae}, we can see in Fig.~\ref{fig:pca-stab-a} and~\ref{fig:pca-stab-b} that when using a \ac{pca} to initialize the \ac{ae} the features look very similar.
They are not identical, but some patterns are shared.
Note that these features are in the third layer; features from the first and second layers, which are not shown for space reason, are almost indistinguishable.

\subsection{Feature Activation Map}
Visualizing features as shown in Fig.~\ref{fig:pca-stability} is not necessarily helpful to know where they are active or not.
For this reason, we generate heatmaps showing where some features are the most active.
We center the \ac{cae} on each pixel of the input image, compute its output, and store the activation value of the three first values as the red, green and blue components of the corresponding pixel of the output image.
Note that there is no labelled training involved in this task -- the \ac{ae} are completely unsupervised.

\begin{figure}[tb]
  \centering
  \subfloat[CS857 sample]{\label{fig:act-base}
    \includegraphics[width=.47\columnwidth]{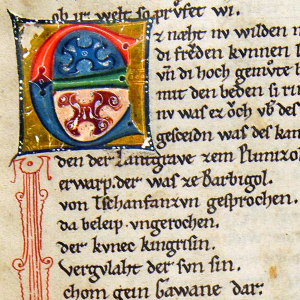}
  }
  \hfill
  \subfloat[PCA activation]{\label{fig:act-pca}
    \includegraphics[width=.47\columnwidth]{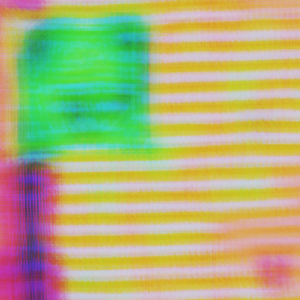}
  } \\
  \subfloat[Xavier activation]{\label{fig:act-xav}
    \includegraphics[width=.47\columnwidth]{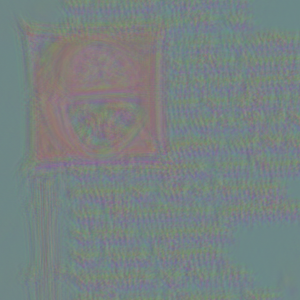}
  }
  \hfill
  \subfloat[Xavier enhanced]{\label{fig:act-xav2}
    \includegraphics[width=.47\columnwidth]{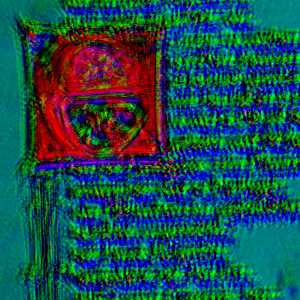}
  }
  \caption{
    Activation of the three first neurons of \ac{cae} on a test image sample.
    The activation values are represented by the intensity of the red, green and blue component of the resulting images.
    As with Xavier initialization this leads to a grayish result, we also show a second version with color enhancement.
  }
  \label{fig:act}
\end{figure}

Very good features should have high levels of correlation with the classes to discriminate.
We can see in Fig.~\ref{fig:act-pca} that the \ac{pca}-initialized \ac{ae} leads to a smooth representation of its input, with clear distinctions between the main part of the decoration (green), the thin red lines below the decoration (purple), and also hints the text line positions (yellow) and interlines (gray).
However, while the Xavier-initialized \ac{ae} seems to capture some areas of the main part of the decoration, the feature activation map is more noisy, and the colors seem less correlated to the kind of content, as there is mostly only red, green and blue.
The background is however better represented by the Xavier-initialized \ac{ae}.

\subsection{Classification Task}
We evaluate our method for a document image layout analysis task at pixel-level.
The goal is to label the pixels of a document image as corresponding to different kinds of content, e.g., background or interlinear glosses.
This is a continuation of our previous work~\cite{chen2015page}, with the main difference that instead of extracting features from the central position of each layer and feed them to an \ac{svm}, we add a classification layer on top of the \ac{ae} and do the backpropagation through the whole network.
This is an approach which we have already successfully applied on the manuscripts of the DIVA-HisDB~\cite{simistira1656a}.

The datasets which we use are split into three parts: training, test and validation.
We used document images from both training and test subsets.
However, as one of our goal is to study the accuracy on the test data during the training, a large quantity of tests have had to be done.
For this reason, we decided to use the following restrictions: we used only two pages from the training set, and test networks on the labelling of 10'000 pixels from the test set.
Also in order to run a fair comparison between Xavier and \ac{pca} initializations, we use the same training and test samples for all networks trained on a same manuscript.
Thus, differences of accuracy at any point in the training are due only to the initial weights of a network.

\begin{figure}[tb]
  \centering
  \subfloat[CS18, short]{\label{fig:classification-short}
    \includegraphics[width=.45\columnwidth]{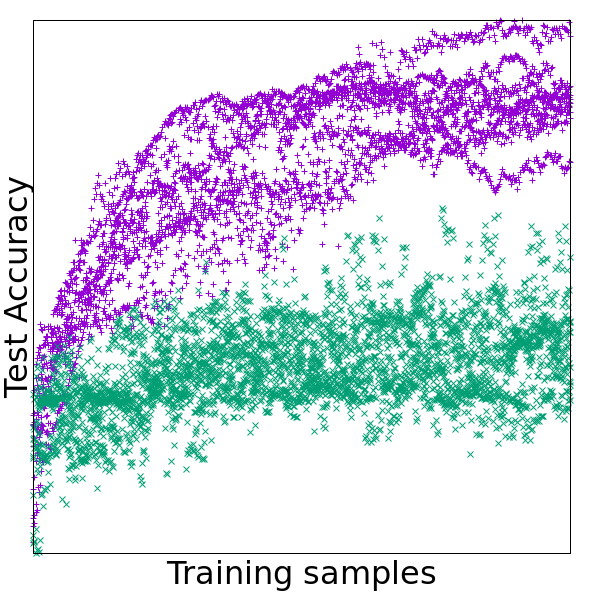}
  }
  \hfill
  \subfloat[CS18, long]{\label{fig:classification-long}
    \includegraphics[width=.45\columnwidth]{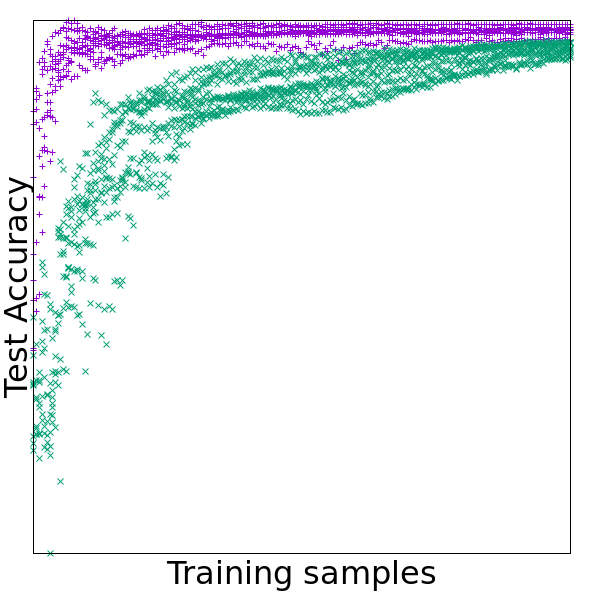}
  } \\
  \subfloat[CB55, long]{\label{fig:classification-long-55}
    \includegraphics[width=.30\columnwidth]{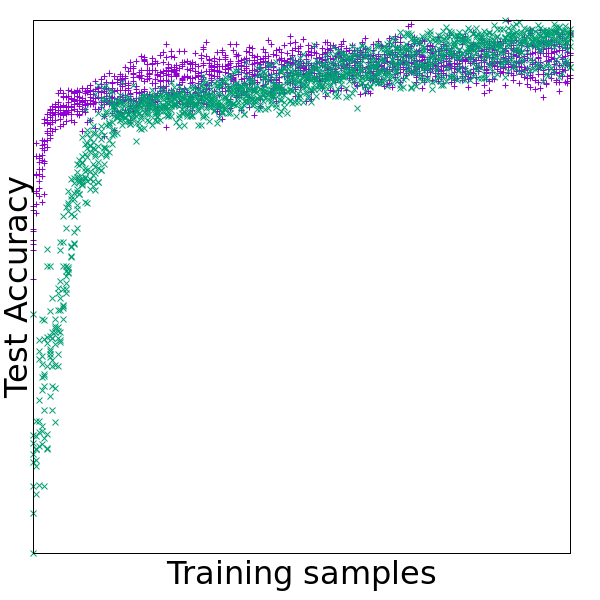}
  }
  \hfill
  \subfloat[GW10, short]{\label{fig:classification-short-10}
    \includegraphics[width=.30\columnwidth]{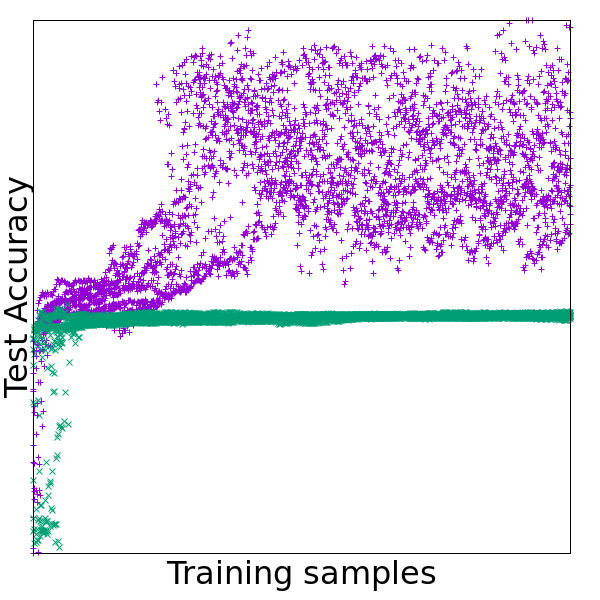}
  }
  \hfill
  \subfloat[GW10, long]{\label{fig:classification-long-10}
    \includegraphics[width=.30\columnwidth]{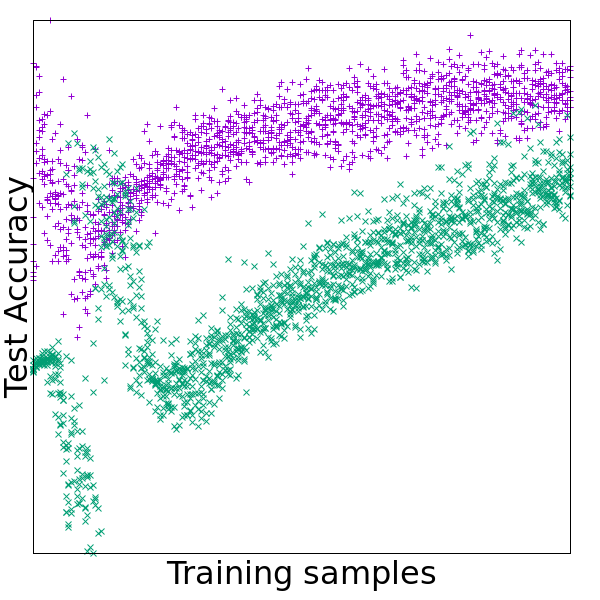}
  } \\
  \subfloat[SG562, long]{\label{fig:classification-long-562}
    \includegraphics[width=.30\columnwidth]{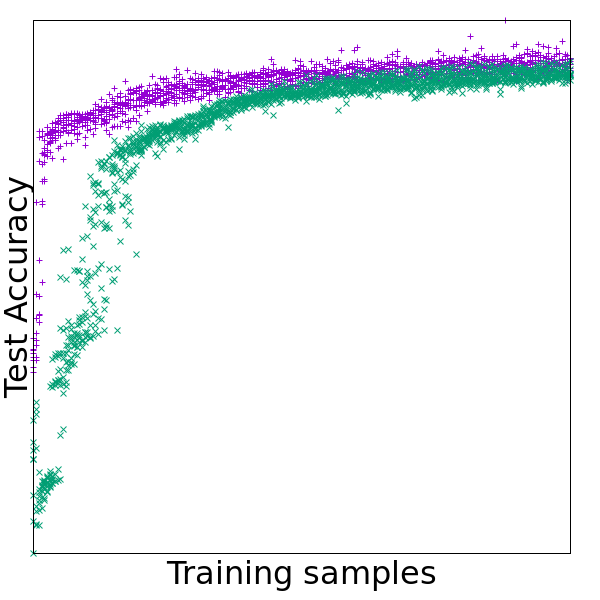}
  }
  \hfill
  \subfloat[CS863, long]{\label{fig:classification-long-863}
    \includegraphics[width=.30\columnwidth]{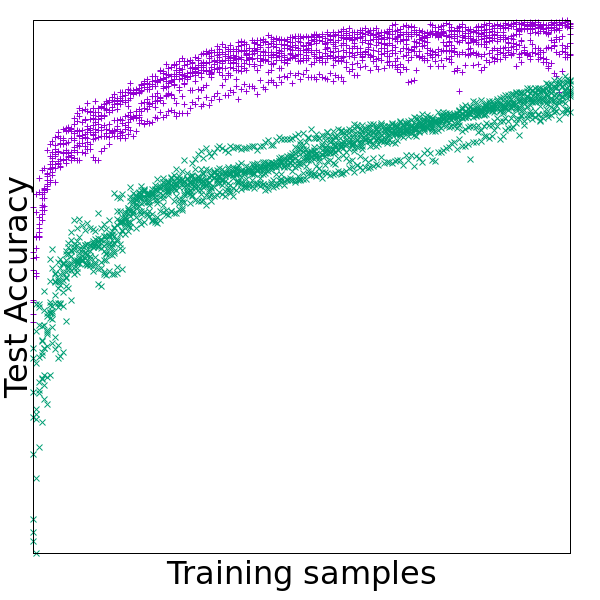}
  }
  \hfill
  \subfloat[SG857, long]{\label{fig:classification-long-857}
    \includegraphics[width=.30\columnwidth]{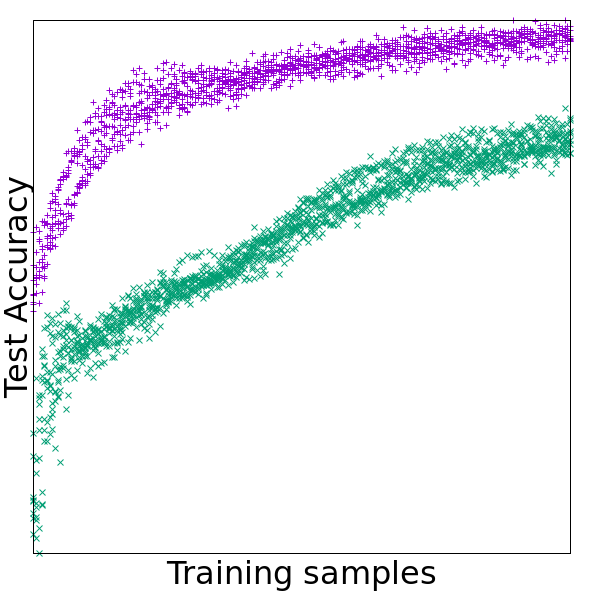}
  }
  \caption{
    Test accuracy of networks during training, for different manuscripts.
    Each image contains data from 16 networks: 8 using Xavier initialization (green), and 8 using \ac{pca} initialization.
    Short training uses 50'000 training samples (1 epoch), and tests were run every 100 samples.
    Long training uses 2'000'000 training samples (1 epoch), and tests were run every 10'000 samples.
  }
  \label{fig:pca-vs-xavier}
\end{figure}

The results obtained during the training are shown in Fig.~\ref{fig:pca-vs-xavier}.
We can see that \ac{pca} initialization always outperforms Xavier initialization, although after a long training they converge toward similar results.
We can however see that although the initial features obtained by \ac{pca} are visually stable, the variance of the accuracy for the two approaches is roughly the same at any point of the training, excepted for the networks trained on CS18 (Fig.~\ref{fig:classification-long}) because the \ac{pca}-initialized networks finished converging very quickly.
This is explained by the fact that small differences in initial conditions can lead to very different results~\cite{poincare1899methodes}, thus although rather similar, the \ac{pca} initialization leads to an error entropy similar to a random, i.e., Xavier, initialization.

\subsection{Error Backpropagation}
We can see in Fig.~\ref{fig:pca-vs-xavier} that \ac{pca}-initialized networks converge very quickly.
To investigate this, we define and use the \ac{rbe}:

\begin{equation}
  r_{be}\left(n\right) = \frac{\sum_i \left|e_{n,in,i}\right|}{\sum_i \left|e_{n,out,i}\right|}
\end{equation}

\noindent where $n$ is a network, $e_{n,in,i}$ is the error of backpropagated to the $i$-th neuron of the first neural layer of $n$, and $e_{n,out,i}$ is the error of the output layer of $n$ with regard to the expected classification result.

The \ac{rbe} is directly linked to how much the weights of a network are modified during a gradient descent.
Thus, it can give hints about how fast a network is learning, and can be used to compare two converging networks, larger \ac{rbe} indicating larger learning rate per sample.
However, the \ac{rbe} value alone cannot predict whether a network converges or not -- a diverging network could even lead to apparently favorable \ac{rbe}.
In this paper, all networks are converging correctly, therefore the \ac{rbe} is a reliable measurement.

\begin{figure}[tb]
  \centering
  \subfloat[CS18, short]{\label{fig:rbe-short}
    \includegraphics[width=.45\columnwidth]{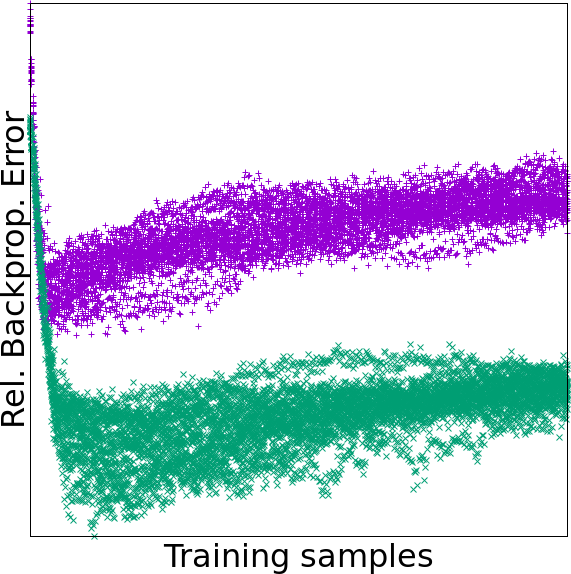}
  }
  \hfill
  \subfloat[CS18, long]{\label{fig:rbe-long}
    \includegraphics[width=.45\columnwidth]{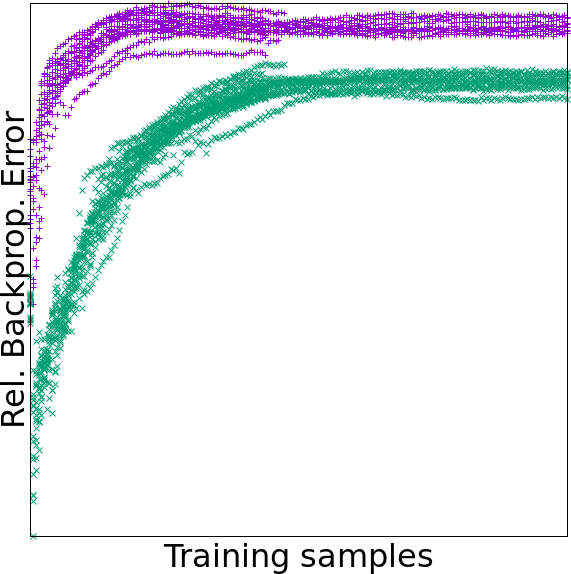}
  } \\
  \subfloat[CB55, long]{\label{fig:rbe-long-55}
    \includegraphics[width=.30\columnwidth]{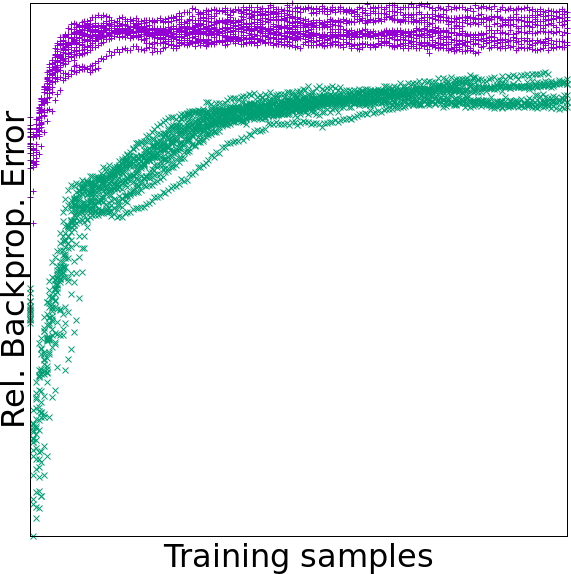}
  }
  \hfill
  \subfloat[GW10, short]{\label{fig:rbe-short-10}
    \includegraphics[width=.30\columnwidth]{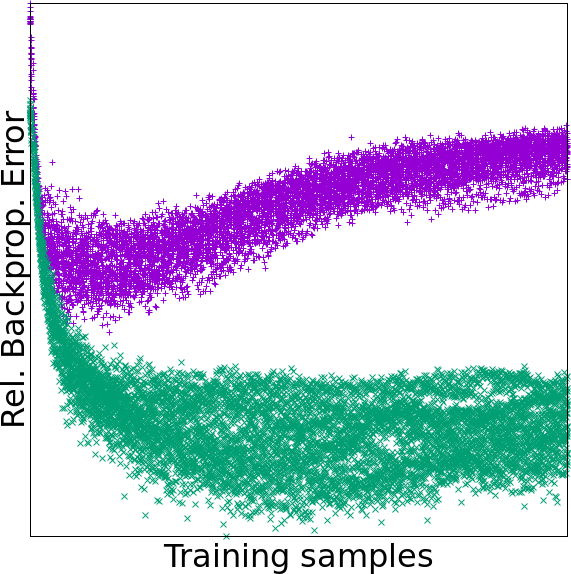}
  }
  \hfill
  \subfloat[GW10, long]{\label{fig:rbe-long-10}
    \includegraphics[width=.30\columnwidth]{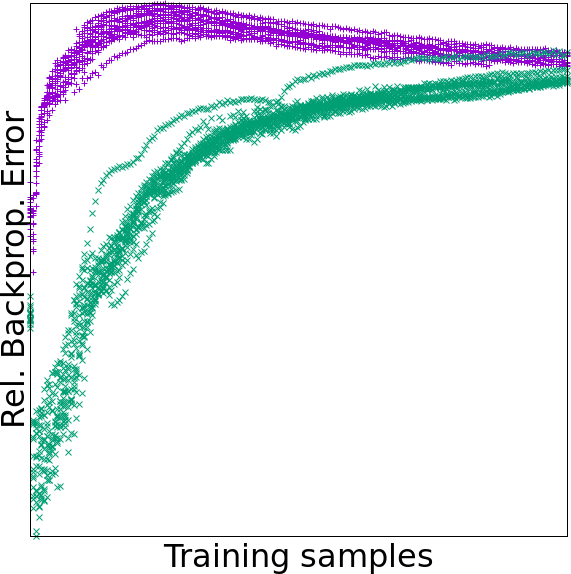}
  } \\
  \subfloat[SG562, long]{\label{fig:rbe-long-562}
    \includegraphics[width=.30\columnwidth]{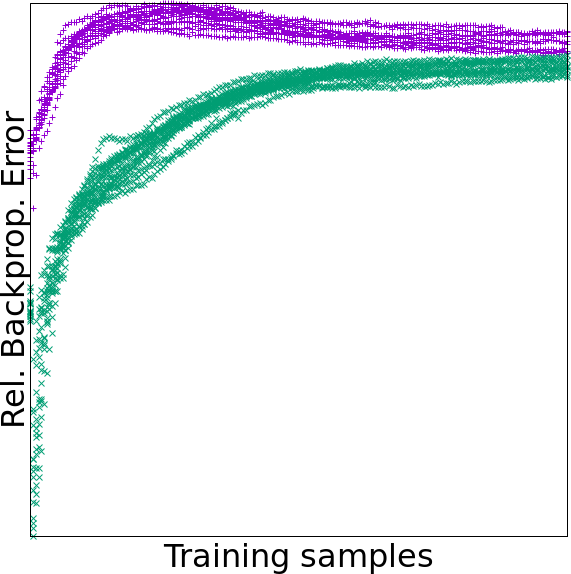}
  }
  \hfill
  \subfloat[CS863, long]{\label{fig:rbe-long-863}
    \includegraphics[width=.30\columnwidth]{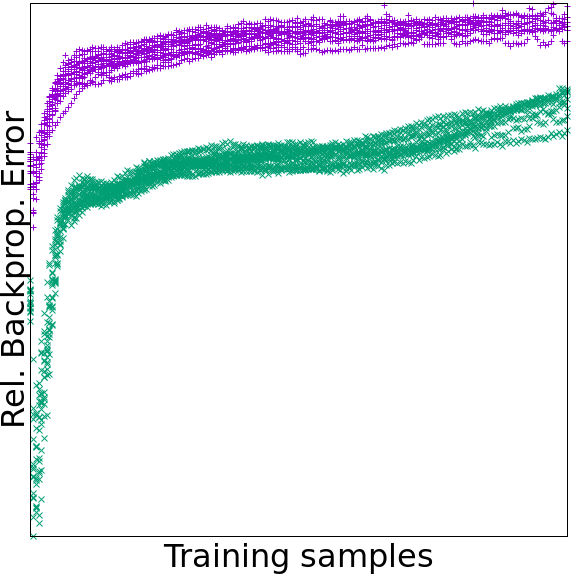}
  }
  \hfill
  \subfloat[SG857, long]{\label{fig:rbe-long-857}
    \includegraphics[width=.30\columnwidth]{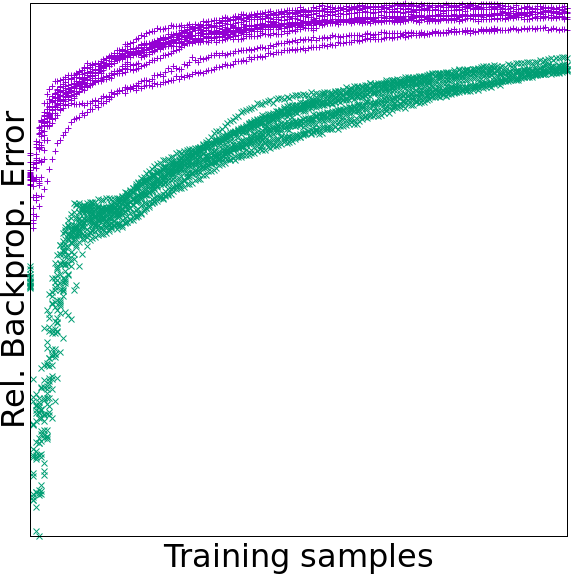}
  }
  \caption{
    \acf{rbe} of networks during training, for different manuscripts.
    Each graphic contains data from 16 networks, where horizontal corresponds to the number of training samples used, and vertical to the logarithm of the sum of \ac{rbe} measured on 100 samples for short training and 10'000 samples for long training.
    Green points correspond to Xavier initialization, and purple points to \ac{pca} initialization.
    Note that the plots match the ones from Fig.~\ref{fig:pca-vs-xavier}.
  }
  \label{fig:rbe}
\end{figure}

We can see in Fig.~\ref{fig:rbe} that \ac{pca}-initialized networks (in purple) always lead to higher \ac{rbe} than Xavier-initialized networks.
This explains why our approach converges much faster.
At the beginning of the training, the \ac{rbe} can be up to 25 times larger when using a \ac{pca} initialization.
After a long training, the \ac{rbe} for both approaches seem to converge toward similar values, in a similar fashion as the accuracy does in Fig.~\ref{fig:pca-vs-xavier}, but even after 2 million weights adjustments, the \ac{pca} initialization still leads to more favorable \ac{rbe}.
This explains why the \ac{pca}-initialized networks always reach a high accuracy earlier than xavier-initialized networks.

\section{Conclusion and Future work}\label{sct:conclusion}
In this paper, we have investigated a new approach for initializing deep auto-encoders using \ac{pca}.
This leads to significantly better performances for layout analysis tasks for an initialization time that can be considered as negligible.
We also investigated the stability of the features learned by \ac{ae} initialized with \ac{pca}, and show that very few samples are required for computing the \ac{pca}.

As future work, we intend to study the use of \ac{lda} to initialize deep auto-encoders for various types of classification tasks.
We assert that this would not only lead to meaningful initializations, as it is the case with \ac{pca}, but also generate weights more useful for classification tasks.

We also hypothesize that a \ac{pca}-initialized network would outperform a Xavier-initialized network in case of transfer learning, as its weights would be tuned faster to the new task. This, however, has to be validated on varios scenarios.

\section*{Acknowledgment}
We would like to thank Kai Chen for feedback he provided during the investigations.
The work presented in this paper has been partially supported by the HisDoc III project funded by the Swiss National Science Foundation with the grant number 205120\_169618.

\bibliographystyle{plain}
\bibliography{bibliography}

\end{document}